\pdfoutput=1

\documentclass[11pt]{article}

\usepackage[final]{acl}

\usepackage{times}
\usepackage{latexsym}

\usepackage[T1]{fontenc}

\usepackage[utf8]{inputenc}

\usepackage{microtype}

\usepackage{inconsolata}

\usepackage{algorithm}
\usepackage{algpseudocode}
\usepackage{multirow}
\usepackage{graphicx}
\usepackage{bbding}
\usepackage{colortbl}
\usepackage{CJKutf8}
\usepackage{booktabs}
\usepackage{amsmath}

\title{Mobile-Agent-V: A Video-Guided Approach for Effortless and Efficient Operational Knowledge Injection in Mobile Automation}

\author{
 \textbf{Junyang Wang\textsuperscript{1}\thanks{Work done during internship at Alibaba Group.}},
 \textbf{Haiyang Xu\textsuperscript{2}\thanks{Corresponding author}},
 \textbf{Xi Zhang\textsuperscript{2}},
 \textbf{Ming Yan\textsuperscript{2}\footnotemark[2]},
\\
 \textbf{Ji Zhang\textsuperscript{2}},
 \textbf{Fei Huang\textsuperscript{2}},
 \textbf{Jitao Sang\textsuperscript{1}\footnotemark[2]},
\\
 \textsuperscript{1}Beijing Jiaotong University,
 \textsuperscript{2}Alibaba Group,
\\
 \small{
   {\{junyangwang, jtsang\}@bjtu.edu.cn}
 }
\\
 \small{
   {\{shuofeng.xhy, ym119608\}@alibaba-inc.com}
 }
}

\begin{document}
\maketitle
\begin{abstract}
The exponential rise in mobile device usage necessitates streamlined automation for effective task management, yet many AI frameworks fall short due to inadequate operational expertise. While manually written knowledge can bridge this gap, it is often burdensome and inefficient. We introduce Mobile-Agent-V, an innovative framework that utilizes video as a guiding tool to effortlessly and efficiently inject operational knowledge into mobile automation processes. By deriving knowledge directly from video content, Mobile-Agent-V eliminates manual intervention, significantly reducing the effort and time required for knowledge acquisition. To rigorously evaluate this approach, we propose Mobile-Knowledge, a benchmark tailored to assess the impact of external knowledge on mobile agent performance. Our experimental findings demonstrate that Mobile-Agent-V enhances performance by 36\% compared to existing methods, underscoring its effortless and efficient advantages in mobile automation.
\end{abstract}

\section{Introduction}

The reliance on mobile devices has increased, with users performing numerous operations daily, underscoring the need for streamlined interactions. Currently, the development of Multimodal Large Language Models (MLLMs) has notably improved mobile device operating frameworks, using these models as intelligent agents \citep{liu2023visual,zhu2023minigpt,ye2023mplug,dai2023instructblip,liu2023improved,chen2023minigpt,bai2023qwen,ye2023mplugowl2,wang2023cogvlm,lu2024deepseek,ye2024mplug,wu2024deepseek,qin2025ui}. These frameworks leverage agents' perception, decision-making, and reflection to perform complex tasks across multiple applications, thereby broadening mobile devices' autonomous capabilities.

\begin{figure}[t]
    \centering
    \includegraphics[width=0.45\textwidth]{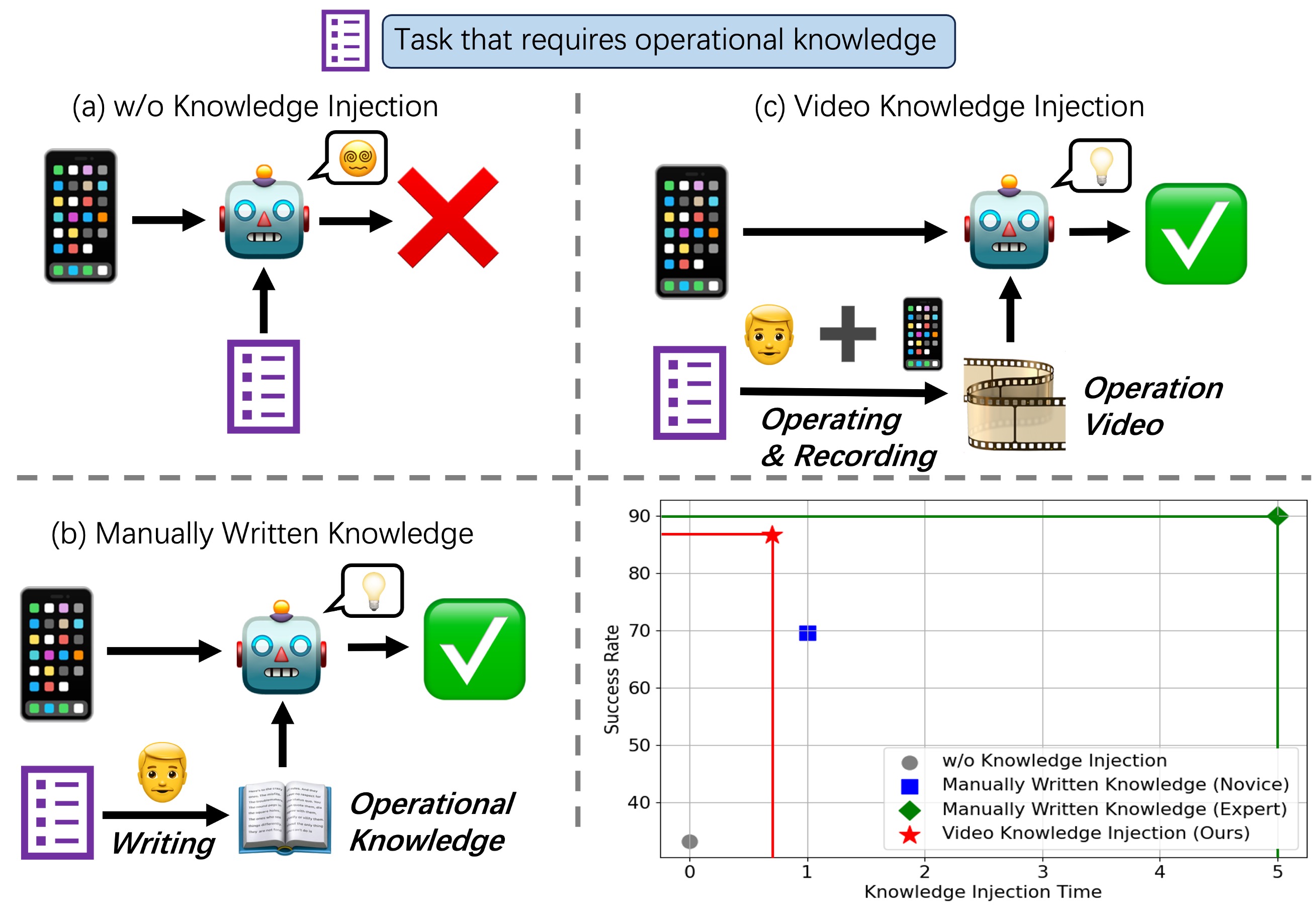}
    \caption{(a) Mobile agents often struggle to complete tasks due to a lack of knowledge. (b) Manually written knowledge requires a high level of human expertise and precision, leading to significant differences in performance depending on whether novices or experts author the content. (c) Mobile-Agent-V learns directly from video, bypassing the need for human expertise. It is more efficient and can even exceed the effectiveness of manually written knowledge. In the evaluation of Mobile-Knowledge, Mobile-Agent-V achieves performance comparable to human experts while saving over 80\% of the time required for knowledge injection.}
    \label{fig:intro}
    \vspace{-3mm}
\end{figure}

Despite progress, existing approaches remain constrained by limited operational knowledge. As shown in Figure~\ref{fig:intro}(a), agents struggle to complete certain tasks when lacking operational knowledge. This is primarily due to the inadequacy of training data to encompass all scenarios. Additionally, the unique nature of some scenarios prevents existing agent knowledge from generalizing effectively. To address this issue, current frameworks typically incorporate manually written knowledge into the agent framework, delivered in textual form \citep{yang2023appagent,li2024appagent,wang2024mobile,wang2024mobile2,agashe2025agent}. However, as depicted in Figure~\ref{fig:intro}(b), this approach is highly sensitive to the quality of human expertise. In order to achieve better outcomes, the involvement of experts becomes necessary. This reliance on manually authored knowledge increases the cost of knowledge injection and reduces efficiency.

To develop methods of knowledge injection that are less reliant on human quality and more efficient, we aim to use knowledge sources in their natural, unprocessed forms. Observations of existing work have shown that video can enhance effectiveness, inspiring us to extract procedural knowledge directly from instructional videos \citep{wang2024videoagent,wang2024lave,zhang2024omagent,chane2023learning}. These videos require users to perform and document an entire operation just once, which removes the need for further human involvement as in Figure~\ref{fig:intro}(c). However, the frequent scene changes and high information density in instructional videos present significant challenges. Additionally, current large-scale visual models often have difficulty processing video input, hindering the ability of existing frameworks to effectively utilize video-based learning.

To address this, we introduce Mobile-Agent-V, a multi-agent framework that processes operational video inputs, extracts actionable knowledge, and applies it to mobile device interactions. To reduce keyframe redundancy while retaining crucial information, we use a sliding window mechanism, feeding a subset of keyframes into the decision agent. The video agent assesses the device's state and adaptively shifts the window forward, ensuring frames remain relevant for decision-making. Despite this, multi-frame inputs challenge MLLMs in maintaining contextual coherence. To enhance accuracy, we employ a reflection agent with long-chain-of-thought reasoning to analyze the video, refine decision outputs.

Existing mobile benchmarks predominantly assess a range of integrated capabilities—such as localization, planning, decision-making, which can conflict with evaluating knowledge utilization, making it difficult to evaluate the effect of knowledge injection alone. To address this, we introduce Mobile-Knowledge, a benchmark designed to specifically assess knowledge utilization efficacy. Utilizing straightforward tasks, it minimizes factors unrelated to knowledge injection. Experimental results indicate Mobile-Agent-V improves performance by 36\% over existing frameworks, demonstrating its superiority in knowledge utilization.

Our summarized contributions are as follows:
\begin{itemize}
    \item We introduce Mobile-Agent-V, a novel framework that applies video guidance to achieve effortless and efficient knowledge injection. Knowledge injection can be accomplished simply by performing the task once and recording a video, eliminating the need for high-quality manual labor and lengthy knowledge construction time.
    \item We propose a multi-agent collaboration strategy to effectively extract and utilize knowledge from videos. To address the challenges of processing long-context video input, we introduce a sliding window strategy in conjunction with a video agent. By incorporating a deep-reflection agent, we further enhance decision accuracy.
    \item To focus on evaluating the effectiveness of knowledge utilization, we introduce Mobile-Knowledge, which comprises tasks that require procedural knowledge but demand minimal basic operational abilities. Experimental results demonstrate that Mobile-Agent-V achieves a 36\% performance improvement over existing frameworks.
\end{itemize}

\begin{figure*}[t]
    \centering
    \includegraphics[width=0.85\textwidth]{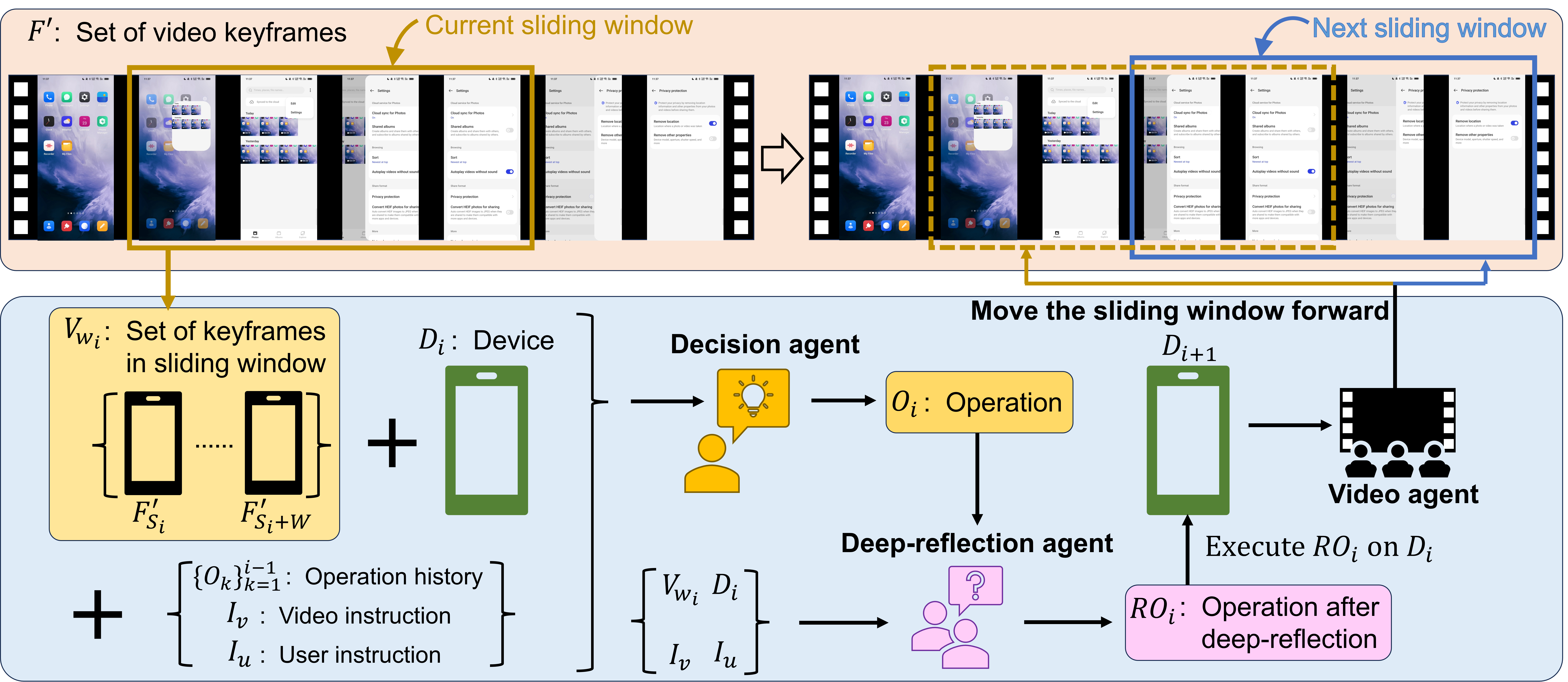}
    \caption{The framework of Mobile-Agent-V.}
    \label{fig:framework}
    \vspace{-3mm}
\end{figure*}

\section{Related Work}

\subsection{GUI Agent}

Intelligent agent frameworks using Large Language Models (LLMs) are advancing in GUI operations to enhance user experience \citep{wang2024gui,liu2025llm}. HTML-based parsing is common on the Web due to its interpretability, while frameworks such as ChatGPT's assistant use visual perception \citep{zhou2023webarena,deng2023mindweb,zheng2024gpt,he2024webvoyager,lu2024weblinx,yoran2024assistantbench,reddy2024infogent}. PC-based frameworks rely on system APIs for greater control \citep{zhang2024ufo,tan2024towards,xie2024osworld}. Mobile automation challenges involve providing agents with operational knowledge, which LLMs often lack. Existing approaches often involve costly training on operational data \citep{hong2023cogagent,cheng2024seeclick,you2024ferret,zhang2024android,chen2024octopus,lu2024gui,chai2024amex,rawles2024androidworld,xu2024androidlab,li2024effects,wan2024omniparser,xing2024understanding,liu2024autoglm}, extensive exploration \citep{yang2023appagent,wang2024mobile,li2024appagent,wang2025mobile}, or inefficiencies through manual knowledge \citep{wang2024mobile2}.

\subsection{Video-guided Agent}

Video guidance is crucial for training intelligent agents to effectively interact with dynamic environments. Initial efforts using large language models (LLMs) focused on video comprehension \citep{wang2024videoagent}. Beyond comprehension, video applications include automated video editing \citep{wang2024lave}, efficient frame retrieval \citep{zhang2024omagent}, and robotics training via human demonstration videos \citep{chane2023learning}. These practical uses showcase the expanding role of video-guided agents in various fields.

\section{Mobile-Agent-V}

This section introduces Mobile-Agent-V, a framework that enhances mobile automation through video guidance. We outline its key components, including video processing, sliding window, video agent, deep-reflection agent, decision agent, and explain how they work together to improve operational efficiency and accuracy.

\subsection{Framework}
The overall workflow of Mobile-Agent-V is shown in Figure~\ref{fig:framework}. Given an input video $V$ that captures a demonstrated task, the system first extracts keyframes $F'$ through uniform sampling and redundancy removal. The execution begins with an initial sliding window positioned at the start of the keyframe sequence. At each iteration, the decision agent generates an action $O_i$ based on the current window, video instructions, and historical decisions. If the task is successfully completed, the process terminates. Otherwise, the deep-reflection agent validates and refines the action to ensure alignment with the demonstrated task. The refined decision $RO_i$ is then executed on the device, updating its state to $D_{i+1}$. The video agent subsequently determines the next window starting point $S_{i+1}$, facilitating a dynamic adjustment of the observation scope as the task progresses. This iterative procedure continues until the task is completed or the predefined maximum exploration limit is reached. The complete pipeline is outlined in Algorithm~\ref{pipeline}.

\subsection{Video Processing}
Traditional uniform sampling suits real-world videos with static scenes and smooth motion. However, in mobile recordings, most frames are static, while rapid changes occur due to human interaction and fast device responses, rendering uniform sampling ineffective for mobile videos. To address this, we first uniformly sample the $V$ at a frequency $d$ to obtain the keyframe set $F$:
\begin{equation}
\label{uniform_sampling}
    F = \text{Uniform\_Sampling}(V, d)
\end{equation}
Next, we compute the similarity between consecutive keyframes and remove those with similarity above a threshold $s$, resulting in a reduced set $F_s$:
\begin{equation}
\label{similarity}
    F_s = \{ f_i \in F \mid \text{sim}(f_i, f_{i+1}) \leq s \}
\end{equation}
Finally, we filter out keyframes with temporal gaps smaller than a threshold $f_s$, yielding the final set of keyframes $F'$:
\begin{equation}
\label{key_frame}
    F' = \{ f_i \in F_s \mid t_{i+1} - t_i \geq d \}
\end{equation}
where $t_i$ represents the frame index of $f_i$.

\subsection{Sliding Window}
To improve video comprehension by MLLMs, we reduce the input length by selecting only the keyframes relevant to the current operation. This is achieved using a sliding window, where the keyframes between the window's start and end points $V_w$ serve as the input for decision-making:
\begin{equation}
\label{window}
    V_w = \{F'_k\}_{k=S_i}^{S_i+W}
\end{equation}
where the $w$ is the length of the window.

\subsection{Decision Agent}
\noindent \textbf{Action Space.} The decision agent is responsible for generating actions that alter the device state. Mobile-Agent-V defines six fundamental actions: \textit{Click}, \textit{Scroll}, \textit{Type}, \textit{Back}, \textit{Home}, and \textit{Done}. A detailed description of the operating space is shown in the Appendix~\ref{action_space_detail}.

\noindent \textbf{Decision Making.} Unlike prior methods that rely on internal operational knowledge, the decision agent in Mobile-Agent-V derives actions directly from video content. This imposes higher demands on contextual adherence. By leveraging the sliding window mechanism, we filter out irrelevant frames, reducing input length while preserving critical information. The $i$-th operation $O_i$ follows the steps outlined in the following equation:
\begin{equation}
    O_i = Da(Vw_i, I_v, D_i, I_u, \{O_k\}_{k=1}^{i-1})
\end{equation}
where $Da(\cdot)$ is the decision agent, $I_v$ is the instruction completed in the video, $D_i$ is the screenshot of the device during the $i$-th operation, and $I_u$ is the instruction that the user will complete on the current device. Besides this, to track the progress, we also provide the historical operations $\{O_k\}_{k=1}^{i-1}$ to the decision agent.

\subsection{Deep-Reflection Agent}
Even with a sliding window, low-quality keyframes require larger window sizes because a smaller window may be filled with redundant frames, excluding important keyframes. In cases where perfect keyframe extraction is not possible, the decision agent struggles with long multi-frame sequences. To overcome this, we introduce the deep-reflection agent, which validates and refines the decision agent’s outputs. It systematically analyzes each operation in the video, identifies the current device state, checks if the decision agent's action matches the corresponding video operation, and refines the action based on the trajectory if discrepancies are found. This reflection mechanism enhances decision accuracy by ensuring strict adherence to the demonstrated operations, leading to a final refined decision $RO_i$, formulated as follows:
\begin{equation}
    RO_i = Ra(Vw_i, I_v, D_i, I_u, O_i)
\end{equation}

\subsection{Video Agent}
To dynamically adjust the sliding window throughout task execution, we introduce the video agent. Initially, the window spans from the first keyframe to the $W$-th keyframe. After each operation, the video agent analyzes the screenshots before and after the operation, keyframes within the current window, and user inputs to identify the corresponding keyframe. Then, it determines the updated window starting point, ensuring adaptive progression. The following is the formula for obtaining the starting point of the $i+1$-th sliding window:
\begin{equation}
    S_{i+1} = Va(Vw_i, I_v, R_i, I_u)
\end{equation}
where $Va(\cdot)$ is the video agent, and $R_i$ is the set of screenshots before and after the operation:
\begin{equation}
    R_i = \{D_k\}_{k=i}^{i+1}
\end{equation}

\begin{algorithm}[t]
\caption{Mobile-Agent-V pipeline}
\label{pipeline}
\begin{flushleft}
\textbf{Input:}
Video $V$, Window length $W$, Video task $I_v$, User task $I_u$, Decision agent $Da$, Reflection agent $Ra$, Video agent $Va$, Max explorations $M_e$\\
\end{flushleft}
\begin{algorithmic}[1]
\State \textbf{Initialization:}
\State Obtain $F'$ from $V$ as Equ.~(\ref{uniform_sampling})~(\ref{similarity})~(\ref{key_frame})
\State $S_1 \gets$ 1
\For{$i = 1$ \textbf{to} $M_e$}
    \State Obtain $V_{w_i}$ from $F'_k$ as Equ.(~\ref{window})
    \State $O_i \gets Da(Vw_i, I_v, D_i, I_u, \{O_k\}_{k=1}^{i-1})$
    \If{$O_i$ == Done}
        \State break
    \EndIf
    \State $RO_i \gets Ra(Vw_i, I_v, D_i, I_u, O_i)$
    \State $D_{i+1} \gets $ Execute $RO_i$ on Device
    \State $R_i \gets \{D_k\}_{k=i}^{i+1}$
    \State $S_{i+1} \gets Va(Vw_i, I_v, R_i, I_u)$
\EndFor
\end{algorithmic}
\end{algorithm}

\section{Experiments}
This section presents a comprehensive evaluation of Mobile-Agent-V. We first introduce the evaluation methodology. Next, we describe the experimental setup. We then report the main results. Finally, we conduct qualitative analyses and ablation studies to further examine the contributions of individual components.

\subsection{Evaluation}
In this subsection, we will introduce the evaluation benchmarks and corresponding metrics.

\subsubsection{Benchmark}
\noindent \textbf{Mobile-Knowledge.} Traditional benchmarks like AITW assess agents' planning and operational skills, including task decomposition, UI element localization, and gesture execution. While these metrics are effective for evaluating basic competencies, they often mix inherent abilities with external knowledge integration. Mobile-Knowledge specifically targets the second dimension. This benchmark minimizes planning and operational complexity, instead emphasizing tasks reliant on knowledge not covered in standard agent training data. We crafted 30 device-specific tasks, categorized as basic, normal, and advanced instructions, each requiring increasing levels of specialized knowledge. Each instruction provides clear directives to avoid biases not related to knowledge integration. For each task, corresponding videos and manually compiled knowledge were provided, with professional annotators supplying the expertise-driven knowledge. Details of the tasks are available in Appendix~\ref{benchmark_detail}.

\noindent \textbf{AndroidWorld-Knowledge.} To evaluate the knowledge generalizability, we developed AndroidWorld-Knowledge within the Android World \citep{rawles2024androidworld} environment. We selected five applications—Expense, Marker, Receipt, SportsTracker, and Tasks—comprising a total of 48 tasks that demand substantial operational knowledge. Within each scenario, only the operation video and manually authored knowledge for the simplest task were provided. This means other tasks in the scenario lacked direct video guidance, relying instead on the least complex task video as a reference. This design assesses the framework's ability to generalize knowledge application beyond direct video instructions. Details of the tasks are available in Appendix~\ref{benchmark_detail2}.

\subsubsection{Metrics}  
We evaluate Mobile-Agent-V and other baselines on Mobile-Knowledge using four key metrics: Success Rate (SR), Completion Rate (CR), Decision Accuracy (DA), and Step Count (Step). The detailed explanation of the evaluation metrics is presented in the Appendix~\ref{metrics}. For AndroidWorld-Knowledge, we follow existing studies by employing SR as a metric to evaluate performance.

\subsection{Setup}
\noindent \textbf{Baselines.} We compare Mobile-Agent-V with several open-source agent frameworks, including AppAgent-v1~\cite{yang2023appagent}, AppAgent-v2~\cite{li2024appagent}, Mobile-Agent-v1~\cite{wang2024mobile}, Mobile-Agent-v2~\cite{wang2024mobile2} and Agent-S2~\cite{agashe2025agent}. For baselines, we utilize manually written knowledge provided by the benchmark for knowledge injection.

\noindent \textbf{Models.} Both Mobile-Agent-V and baselines utilize GPT-4o as their base model. The model is accessed via the official API with default hyperparameters.

\noindent \textbf{Device and Interaction.} Experiments on Mobile-Knowledge are conducted on a OnePlus 7 Pro smartphone using the Android Debug Bridge (ADB) for interaction.

\begin{table*}[t]
	\centering
	\renewcommand{\arraystretch}{1}
	\setlength{\tabcolsep}{10pt}
	\scalebox{0.9}{
	\begin{tabular}{l|c|c c c c}
        \hline
		\toprule
	Method&Knowledge Injection&SR&CR&DA&Step\\
        \midrule
        AppAgent-v1 \citep{yang2023appagent}&Manually Written&46.7&52.5&43.6&12.2\\
        AppAgent-v2 \citep{li2024appagent}&Manually Written&60.0&67.3&57.7&10.8\\
        Mobile-Agent-v1 \citep{wang2024mobile}&Manually Written&43.4&51.3&41.0&12.2\\
        Mobile-Agent-v2 \citep{wang2024mobile2}&Manually Written&56.6&59.8&54.8&11.4\\
        Agent-S2 \citep{agashe2025agent}&Manually Written&63.3&73.9&60.1&13.6\\
        \midrule
        Mobile-Agent-V (Ours)&Operation Video&\textbf{86.7}&\textbf{93.4}&\textbf{79.4}&\textbf{7.3}\\
        \bottomrule
        \hline
	\end{tabular}
	}
    \caption{Evaluation results on Mobile-Knowledge benchmark.}
	\label{tb:mobile-knowledge}
    \vspace{-3mm}
\end{table*}

\begin{table}[!ht]
	\centering
	\renewcommand{\arraystretch}{1}
	\setlength{\tabcolsep}{8pt}
	\scalebox{0.9}{
	\begin{tabular}{l|c}
        \hline
		\toprule
	Method&SR\\
        \midrule
        AppAgent-v1 \citep{yang2023appagent}&14.6\\
        AppAgent-v2 \citep{li2024appagent}&18.9\\
        Mobile-Agent-v1 \citep{wang2024mobile}&12.5\\
        Mobile-Agent-v2 \citep{wang2024mobile2}&16.7\\
        Agent-S2 \citep{agashe2025agent}&18.9\\
        \midrule
        Mobile-Agent-V (Ours)&\textbf{31.3}\\
        \bottomrule
        \hline
	\end{tabular}
	}
    \caption{Evaluation results on AndroidWorld-Knowledge benchmark.}
	\label{tb:andriodworld-knowledge}
    \vspace{-3mm}
\end{table}

\subsection{Main Results}
In this subsection, we will analyze the performance of different methods on the Mobile-Knowledge and AndroidWorld-Knowledge benchmarks.

\subsubsection{Mobile-Knowledge}
The results on the Mobile-Knowledge benchmark highlight the effectiveness of Mobile-Agent-V, which utilizes operation video for knowledge injection. Compared to baseline methods that rely on manually written knowledge, Mobile-Agent-V shows a significant improvement in metrics such as SR, CR, and DA, with enhancements of up to 23.4\% over the best-performing baseline. Additionally, Mobile-Agent-V achieves greater action efficiency, as evidenced by a reduction in the Step metric. These outcomes underscore the advantages of integrating operation videos, offering a more dynamic and comprehensive understanding of tasks than static instructional text.

\subsubsection{AndroidWorld-Knowledge}
On the AndroidWorld-Knowledge benchmark, Mobile-Agent-V demonstrates a substantial improvement in SR over baselines, achieving a 31.3\% SR. This represents a significant increase of at least 12.4\% compared to the best baseline, highlighting the effectiveness of utilizing operation videos for knowledge integration. The notable performance gain emphasizes Mobile-Agent-V's capability to enhance generalizability and operational efficiency in diverse GUI tasks, surpassing traditional approaches that depend solely on manually written instructions. Since AndroidWorld-Knowledge provides only one video per scenario, it facilitates the evaluation of generalization when discrepancies arise between the operation video and the actual task. We will conduct a detailed analysis of the generalization derived from video knowledge in Section~\ref{generalization}.

\subsection{Analysis}
We conducted analytical experiments on the framework's configuration using the Mobile-Knowledge.

\begin{figure*}[t]
    \centering
    \includegraphics[width=0.975\textwidth]{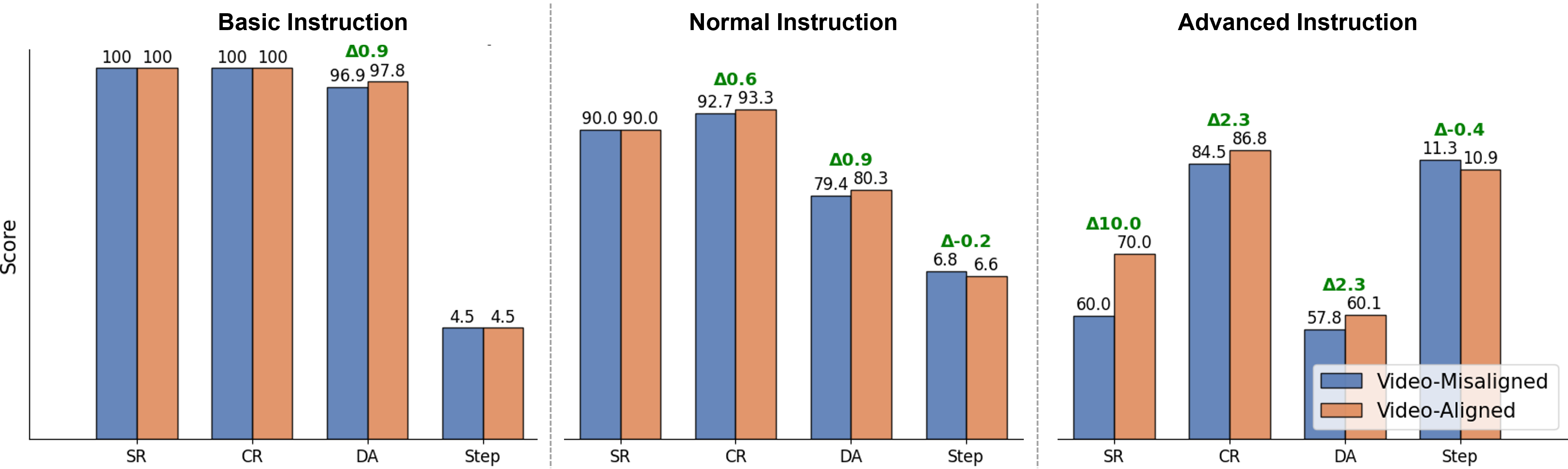}
    \caption{Comparison of video-misaligned instructions and video-aligned instructions. The video-aligned means that the video instruction is consistent with the user instruction, and the video-misaligned instruction is inconsistent.}
    \label{fig:CD}
    \vspace{-3mm}
\end{figure*}

\begin{figure*}[t]
    \centering
    \includegraphics[width=0.95\textwidth]{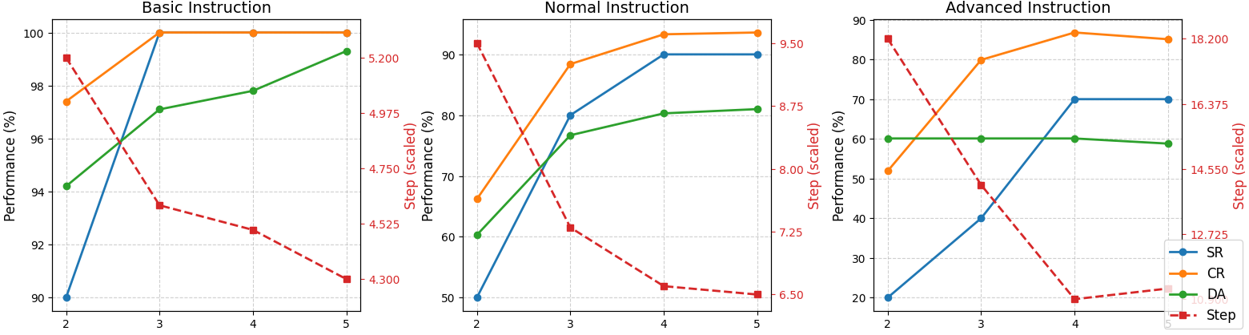}
    \caption{Comparison of different sliding window sizes.}
    \label{fig:window}
    \vspace{-3mm}
\end{figure*}

\begin{figure}[!ht]
    \centering
    \includegraphics[width=0.35\textwidth]{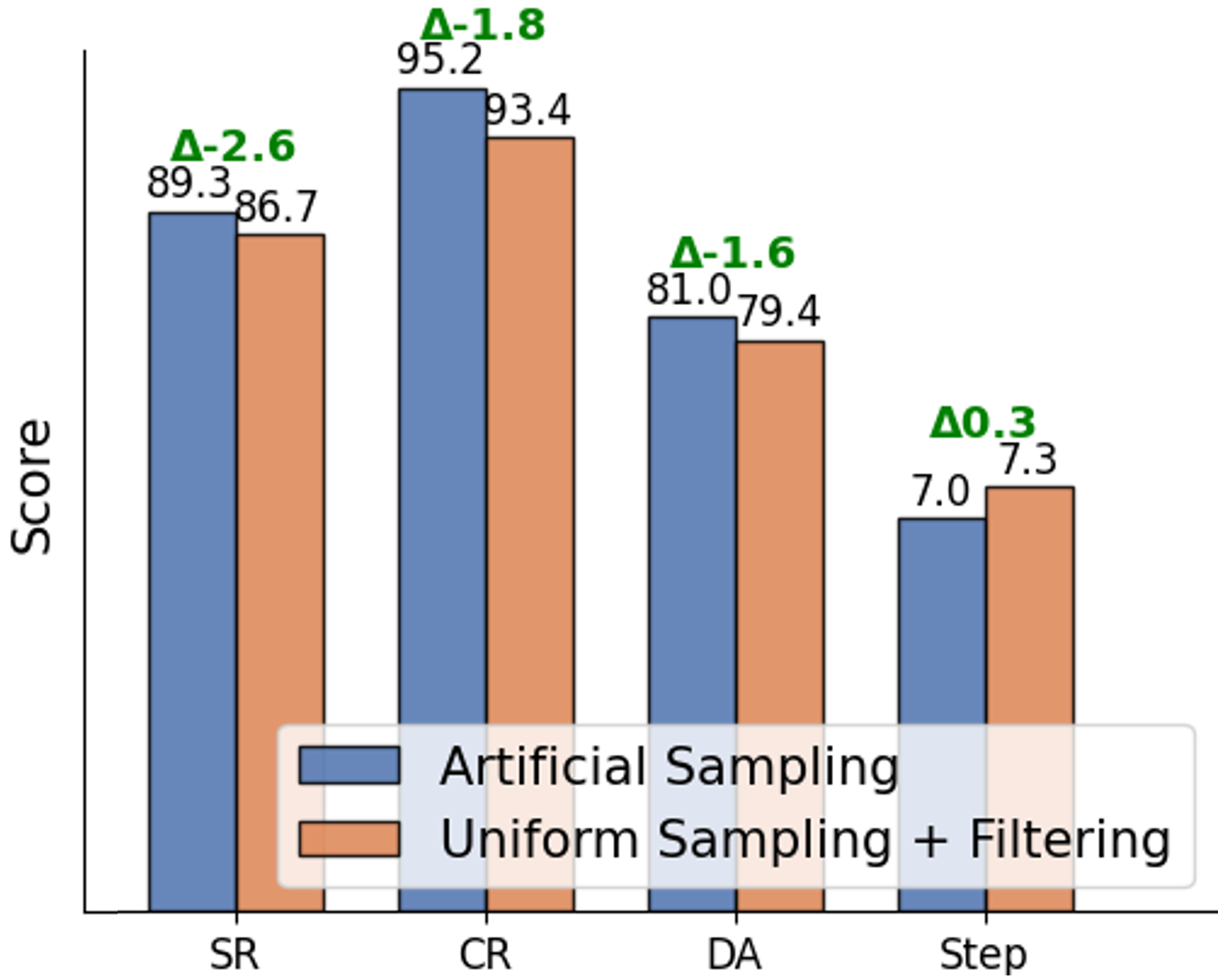}
    \caption{Comparison of different keyframe quality.}
    \label{fig:keyframe}
    \vspace{-3mm}
\end{figure}

\begin{figure*}[t]
    \centering
    \includegraphics[width=0.95\textwidth]{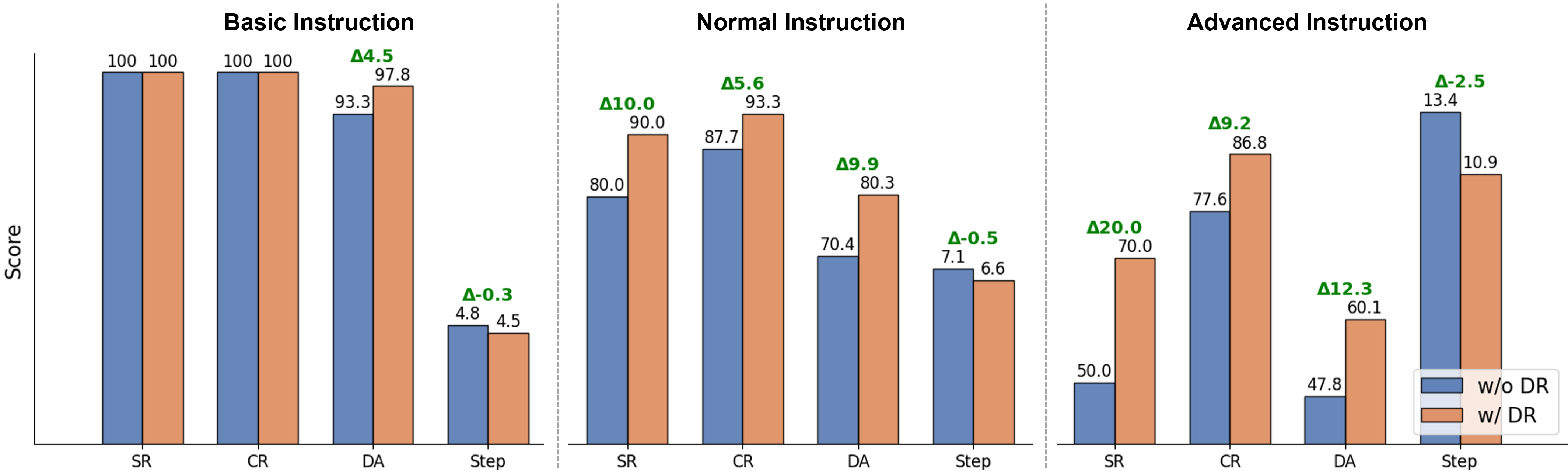}
    \caption{Comparison of w/o DR and w/ DR across different instructions.}
    \label{fig:DR}
\end{figure*}

\begin{figure*}[!ht]
    \centering
    \includegraphics[width=0.88\textwidth]{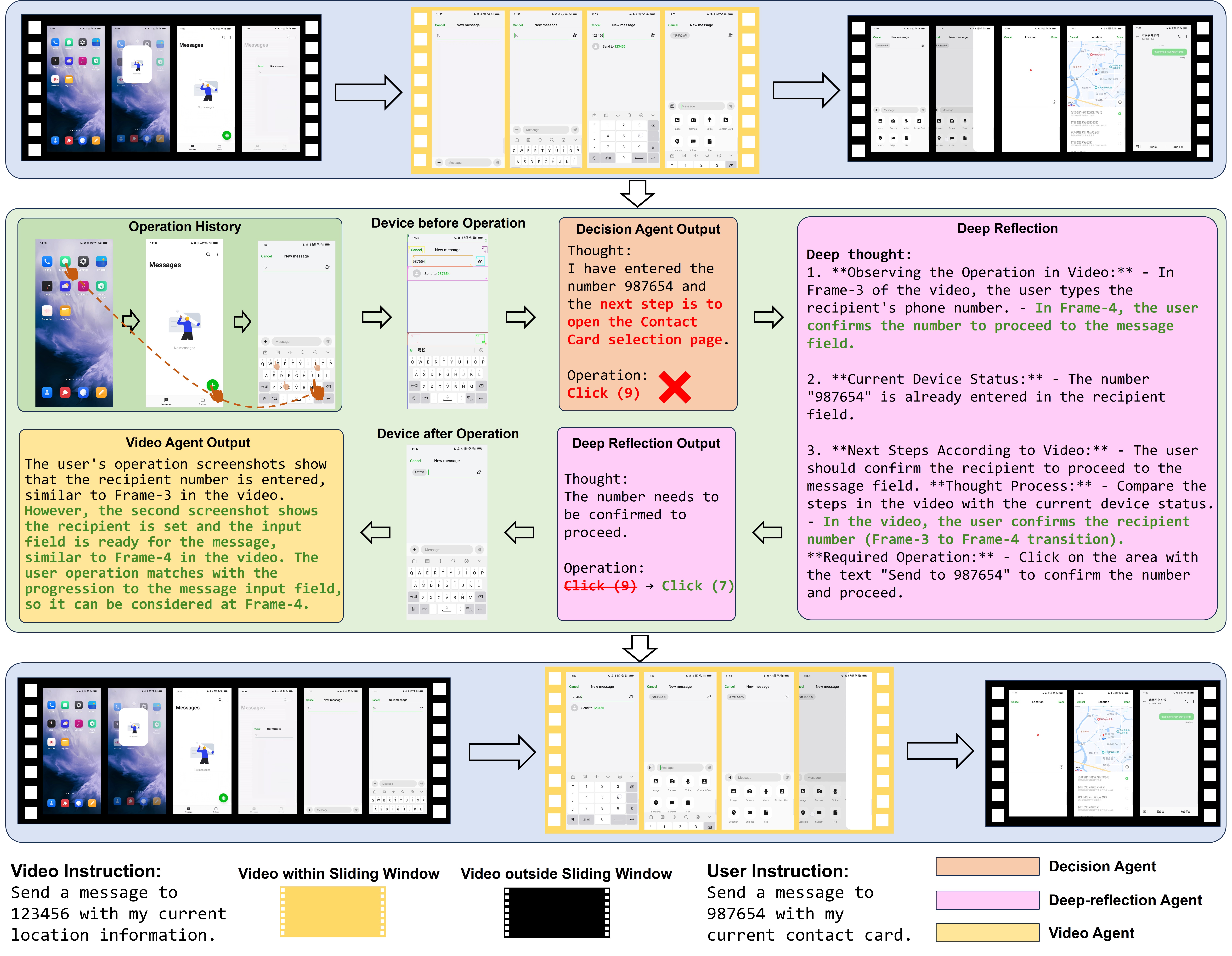}
    \caption{A complete execution case of Mobile-Agent-V. The decision agent initially makes an incorrect action, but the deep-reflection agent verifies the operation video, compares the device state, and corrects the action.}
    \label{fig:case}
    \vspace{-3mm}
\end{figure*}

\subsubsection{Generalization from Videos}
\label{generalization}
The Video-Misaligned task modifies original instructions so the video’s operational logic aligns with the user task, but actions differ. This tests Mobile-Agent-V's ability to generalize from video demonstrations. As shown in Figure~\ref{fig:CD}, Mobile-Agent-V's performance drops under Video-Misaligned conditions; basic instructions stay stable, while normal and advanced ones decline in SR and DA. Yet, the system still completes tasks competently, indicating its ability to generalize beyond direct instruction mapping. These results emphasize the importance of diverse video demonstrations for enhancing cross-instruction generalization.

Mobile-Agent-V's ability to generalize from videos is a key strength demonstrated on the AndroidWorld-Knowledge benchmark. In this benchmark, we provided only a single video or manually written knowledge for the simplest task in each of the five scenarios. As shown in Table~\ref{tb:andriodworld-knowledge}, despite the potential discrepancies between the provided videos and the actual tasks, Mobile-Agent-V achieved a SR of 31.3\%, significantly outperforming baselines. This indicates that Mobile-Agent-V can effectively extrapolate from limited video input, generalizing to more complex tasks without direct video guidance. This capability underscores the adaptability and robustness of our video-guided approach, which is essential for practical mobile automation applications where task-specific video resources may be limited or unavailable.

\subsubsection{Impact of Window Size}
Figure~\ref{fig:window} illustrates the effect of window size on task performance. Larger windows generally improve SR, CR, and DA while reducing steps, particularly for more complex tasks. However, beyond a certain threshold, further increasing the window size yields diminishing returns, with some metrics even declining. This decline is likely due to the introduction of irrelevant information, which interferes with decision-making. These findings highlight the importance of balancing temporal context to maximize efficiency.

\subsubsection{Impact of Keyframe Quality}
To investigate the impact of keyframe quality, we compare artificial sampling, where keyframes are manually selected to avoid redundancy and omission, with our uniform sampling and filtering strategy in Figure~\ref{fig:keyframe}. As expected, manually chosen keyframes yield slightly better results, confirming that high-quality keyframes enhance performance. However, the gap between our method and manual selection remains small, demonstrating the effectiveness of our method in preserving essential task-relevant information.

\subsubsection{Impact of Knowledge Injection Method}
Figure~\ref{tab:time} highlights the considerable impact of the knowledge injection method on performance and efficiency. Mobile-Agent-V utilizes operation videos, achieving a high SR of 86.7\% while reducing knowledge injection time to just 0.7 minutes on average. It balances the benefits of novice and expert-level manually written knowledge, which, despite higher SRs, require substantial time—up to five minutes for expert knowledge. The efficiency of video-based knowledge aligns with Mobile-Agent-V's goals, focusing on seamless, efficient integration in mobile automation. Mobile-Agent-V provides an optimal solution, enhancing accessibility without sacrificing performance and avoiding the resource-intensive process of manual expertise.

\begin{table}[!ht]
    \centering
    \renewcommand{\arraystretch}{1}
    \scalebox{0.85}{
    \begin{tabular}{l|c c}
        \hline
        \toprule
        \textbf{Knowledge Injection Method} & \textbf{SR} &\textbf{Avg. Time} \\
        \midrule
        - & 33.3 & - \\
        Manually Written - Novice & 70.0 & 1 min \\
        Manually Written - Expert & 90.0 &5 mins\\
        Operation Video & 86.7 & 0.7 min\\
        \bottomrule
        \hline
    \end{tabular}
    }
    \caption{A comparison of the knowledge injection time and performance between video and manually written knowledge across varying levels of human expertise.}
    \label{tab:time}
    \vspace{-3mm}
\end{table}

\subsection{Ablation Study}
To evaluate the deep-reflection agent's effectiveness, we conducted an ablation study comparing its performance with and without the agent, as depicted in Figure~\ref{fig:DR}. Results show that the deep-reflection agent consistently enhances decision-making across metrics. When SR and CR are high, improvements are minor due to fewer errors by the decision agent. However, for complex tasks with lower baseline performance, the deep-reflection agent significantly boosts DA, refining actions and reducing inconsistencies in extended multi-frame reasoning. The Step metric shows slight changes, suggesting improved precision without major impacts on action efficiency. By correcting misalignments between predicted and actual actions, the agent mitigates cascading errors in long-horizon tasks, reduces reliance on perfect keyframe extraction, and enhances robustness and reliability in challenging visual conditions.

\subsection{Case Study}
Figure~\ref{fig:case} presents a multi-agent collaboration scenario within Mobile-Agent-V. The decision agent analyzes keyframes from a sliding window to determine the operation but mistakenly skips the "confirm contact" step, highlighting multi-image action tracking challenges. The deep-reflection agent corrects this by identifying the misalignment and refining the decision to ensure accurate device operation. Meanwhile, the video agent anchors the device state to the fourth frame, then advances the window by two frames, allowing the system to accurately display the next interaction with the contact card.

\section{Conclusion}
We present Mobile-Agent-V, a video-guided framework that advances mobile automation by integrating dynamic, cost-effective operational knowledge. Using a sliding window mechanism, the video agent optimally selects keyframes, while the deep-reflection agent enhances decision accuracy through iterative reasoning. Experiments indicate Mobile-Agent-V's superior performance, with a 23.4\% Success Rate improvement on Mobile-Knowledge and 12.4\% on AndroidWorld-Knowledge. Mobile-Agent-V rivals expert-level written knowledge, reducing injection time by 86\%, underscoring its potential for scalable learning. Mobile-Agent-V effectively transforms videos into operational knowledge, offering a streamlined path for agent development.

\section{Limitations}
While our method offers significant advantages, there are certain limitations to consider. Firstly, the dependency on video inputs may introduce variability in data quality; suboptimal recordings could impact the accuracy of knowledge extraction. Although the sliding window mechanism significantly enhances processing efficiency, there remains a possibility that essential frames could be overlooked during complex interactions. Furthermore, while our framework successfully generalizes across diverse tasks, its performance is somewhat contingent on the range and quality of video demonstrations available. Future work could focus on developing adaptive mechanisms to further improve both the efficiency and robustness of the system, ensuring it can handle a wider array of scenarios with varying video quality.

\bibliography{reference}

\clearpage

\appendix

\section{Appendix}
\label{sec:appendix}

\subsection{Experimental Details}

\begin{table*}[h]
    \centering
    \renewcommand{\arraystretch}{1.2}
    \begin{tabular}{l|l|p{11cm}}
        \hline
		\toprule
        \textbf{Action} & \textbf{Parameter} & \textbf{Description} \\
        \hline
		\toprule
        Click & id & The "id" represents the numeric identifier of the detection box to be clicked. \\
        \hline
        Click\_text & text & The "text" specifies the target text to be clicked, used only when no detection box or corresponding ID exists at the target location. \\
        \hline
        Scroll & direction & The "direction" can be either "up" or "down," allowing the agent to scroll the screen accordingly. \\
        \hline
        Type & text & The "text" parameter defines the content to be entered into a text field. \\
        \hline
        Back & None & Returns to the previous screen. \\
        \hline
        Home & None & Navigates to the home screen. \\
        \hline
        Done & None & Signals task completion. \\
        \bottomrule
        \hline
    \end{tabular}
    \caption{Action space definition for Mobile-Agent-V.}
    \label{action_space}
\end{table*}

This section provides additional details regarding the experimental setup and implementation choices used in Mobile-Agent-V.

\subsubsection{Sliding Window Size Selection}
In our experiments, the sliding window size was set to 4. While increasing the window size to 5 is also feasible, experimental analysis demonstrated that the performance improvement was marginal, while the computational cost increased due to the higher token consumption. Therefore, we adopted a window size of 4 as a balanced trade-off between efficiency and performance.

\subsubsection{Video Similarity Computation}
To compute the similarity between video frames, we employed a simple yet effective approach based on pixel-wise differences. Given two frames $I_1$ and $I_2$, we first converted them to grayscale representations:
\begin{equation}
I'_1 = \text{grayscale}(I_1), \quad I'_2 = \text{grayscale}(I_2)
\end{equation}
Next, we computed the absolute difference between the two grayscale images:
\begin{equation}
D = \text{absdiff}(I'_1, I'_2)
\end{equation}
Finally, the similarity score $S$ was obtained by counting the number of nonzero pixels in $D$:
\begin{equation}
S = \frac{\text{np.count\_nonzero}(D)}{\text{total pixels}}
\end{equation}
This method effectively captures differences between frames while maintaining computational efficiency.

\begin{table*}[t]
	\centering
	\renewcommand{\arraystretch}{1.2}
	\setlength{\tabcolsep}{8pt}
	\scalebox{0.95}{
	\begin{tabular}{p{15cm}}
        \hline
		\toprule
  \textbf{System}\\
  \hline
    You are an expert in mobile phone operation. I will upload two images below. The first image is a keyframe mosaic from an operation video, in which the completed task is "\{$I_v$\}"; the second image is a screenshot of the current status of the mobile phone.\\
    \\
    On the mobile phone shown in the second image, the task to be completed is: "\{$I_u$\}". The user will perform the following operation:\\
    \{Operation from decision agent\}\\
    \\
    Now please observe whether this operation conforms to the operation path shown in the first image. If it conforms, please output "True", otherwise please modify the operation content according to the above json format.\\
    \\
    The operation should be:\\
    - Click (id): The "id" is the numeric serial number of the detection box you need to click.\\
    - Click\_text (text): The "text" is the text you need to click. This is only used when the detection box and the corresponding id do not exist at the location to be clicked.\\
    - Scroll (direction): The "direction" selects from "up" and "down". You can scroll the page a certain distance in the specified direction.\\
    - Type (text): The "text" is the content you need to enter.\\
    - Back: You can use this operation to return to the previous page.\\
    - Home: You can use this operation to return to the home page.\\
    - Done: You can use this operation when the task is completed.\\
    Note: If the operation history and current device can infer that the task has been completed, use Done.\\
    \\
    You need to think in the following way:\\
    1. Observe the operation of each step in the video (especially frame-3 and frame-4).\\
    2. Anchor the position of the current device in the video.\\
    3. Complete the current step according to the operation in the video.\\
    Please output your thought about this step by step before you output your response.\\
  \midrule
    \textbf{User}\\
    \hline
    <image: $V_w$><image: $D_i$>\\
    \bottomrule
    \hline
	\end{tabular}
	}
    \caption{The prompt for deep-reflection agent.}
	\label{tb:prompt_reflection}
\end{table*}

\subsubsection{Frame Similarity Threshold Selection}
As described in the main text, the similarity threshold $f_s$ was adjusted according to the characteristics of different applications. For instance, in the \textit{Settings} app, where UI changes are primarily text-based, we set $f_s=0.3$ to ensure that more informative frames were retained. Conversely, for the \textit{Weather} app, where UI elements exhibit significant visual variations, a higher threshold of $f_s=0.5$ was used to prevent excessive redundant frame extraction.

\subsubsection{Step Limitations and Task Termination Criteria}
To ensure fair evaluation and prevent infinite loops, we imposed an upper bound on the number of execution steps:
\begin{itemize}
    \item Basic tasks: 10-step limit.
    \item Standard tasks: 15-step limit.
    \item Complex tasks: 20-step limit.
\end{itemize}
If an agent reached the step limit without successfully completing the task, the attempt was deemed a failure. Additionally, if a framework executed the required action but continued performing unnecessary operations beyond the instruction’s scope, it was also considered a failure.

\subsubsection{Video Frame Concatenation for Visualization}

To simplify interpretation, video frames were concatenated in a row-wise manner. Each frame within the sliding window was indexed to aid the video agent in tracking its progress. In instances where fewer than four frames were available, only the existing frames (up to three) were concatenated. The final frame in each sequence was distinctly marked as the termination state, guiding the decision agent to stop at the correct point.

\subsubsection{Action Space Definition}
\label{action_space_detail}

Mobile-Agent-V utilizes the same action space as Mobile-Agent-V2. Unlike Mobile-Agent-V2, which employs OCR and segmentation models to identify interaction coordinates, Mobile-Agent-V uses the Set of Mark (SoM) approach to decrease context length. To address potential XML parsing issues in certain UI pages, a supplementary click-by-text operation was introduced. A complete outline of the action space is provided in Table~\ref{action_space}.

\subsection{Prompt}

Tables~\ref{tb:prompt_reflection}, \ref{tb:prompt_decision}, and \ref{tb:prompt_video} display the prompts used by the deep-reflection agent, decision agent, and video agent, respectively.

\subsection{Benchmark Details}

\subsubsection{Evaluation Tasks of Mobile-Knowledge}
\label{benchmark_detail}

Table~\ref{benchmark} presents a comprehensive breakdown of benchmark tasks, categorized by application. This structure evaluates Mobile-Agent-V’s proficiency in interpreting, aligning, and executing user instructions of varying complexity. The benchmark differentiates between video-aligned and video-misaligned instructions, testing the framework’s robustness against linguistic variations and its adaptability to real-world user interactions.

\subsubsection{Evaluation Tasks of AndroidWorld-Knowledge}
\label{benchmark_detail2}

Table~\ref{benchmark2} shows the task names from Android World in AndroidWorld-Knowledge.

\subsubsection{Metrics}
\label{metrics}

The following metrics characterize the evaluation process:

\begin{itemize}
\item \textbf{Success Rate}: This metric represents the percentage of instructions that are fully completed, offering a comprehensive measure of the agent's capability in executing tasks from start to finish without errors. A high success rate indicates proficient end-to-end execution, underscoring the agent’s overall effectiveness and reliability in automating tasks accurately and efficiently.

\item \textbf{Completion Rate}: Completion Rate quantifies the proportion of individual steps executed within a given instruction, providing a more granular view of task progression. This metric is essential for understanding areas where the agent may excel or face challenges, particularly in the execution of sequential tasks. By analyzing completion rates, researchers and developers can identify specific steps that require optimization or redesign to enhance overall task completion.

\item \textbf{Decision Accuracy}: This metric evaluates the precision of the agent's decision-making processes by comparing the number of correctly made decisions against the total number of decisions attempted. High decision accuracy reflects the agent’s adeptness in selecting appropriate actions based on provided data, highlighting its ability to navigate complex decision spaces effectively.

\item \textbf{Step Count}: Step Count provides insight into the number of actions the agent takes to accomplish a given instruction and acts as a measure of execution efficiency. By tracking the steps required for task completion, this metric aids in pinpointing inefficiencies and excessive actions that may hinder performance.
\end{itemize}

\subsubsection{Screen Recording}

All videos were captured using the built-in screen recording tool on a OnePlus 7 Pro test device. While the tool supports a maximum frame rate of 60 Hz, practical frame rates ranged between 30 Hz and 60 Hz, contingent upon the degree of UI changes. Interactions were manually performed at an average frequency of one action every 1–2 seconds. The videos were left unprocessed, free from edits such as acceleration or overlays, thus preserving their original state. Each benchmark instruction corresponds to a unique operation video, demonstrating the optimal path for task execution.

\begin{table*}[t]
	\centering
	\renewcommand{\arraystretch}{1}
	\setlength{\tabcolsep}{8pt}
	\scalebox{0.9}{
	\begin{tabular}{p{17cm}}
        \hline
		\toprule
  \textbf{System}\\
  \hline
    You are a mobile phone operation assistant. Below is a description of this conversation.\\
    \\
    In the following part, I will upload a large image made up of many screenshots. These screenshots in this image are all from a screen recording of a mobile phone operation. I will tell you the task completed in the screen recording. You need to observe this screen recording.\\
    \\
    Then, you need to complete a new task, which is related to the task in the screen recording. You need to combine the operation experience provided by the screen recording and gradually complete this task. I will upload the current screenshot of the device. There will be many detection boxes on this screenshot, and there will be a number in the upper left and lower right corners of the detection box. You need to perform operations on the current page. In order to better operate the phone, the following are the operation tools you can use:\\
    - Click (id): The "id" is the numeric serial number of the detection box you need to click.\\
    - Click\_text (text): The "text" is the text you need to click. This is only used when the detection box and the corresponding id do not exist at the location to be clicked.\\
    - Scroll (direction): The "direction" selects from "up", "down", "left", and "right". You can scroll the page a certain distance in the specified direction.\\
    - Type (text): The "text" is the content you need to enter.\\
    - Back: You can use this operation to return to the previous page.\\
    - Home: You can use this operation to return to the home page.\\
    - Done: You can use this operation when the task is completed.\\
    \\
    You need to strictly follow the following json output format:\\
    {"Thought": You need to think about how to perform this operation on the current device based on the operation path in the video, "Operation": Select one from the operation tools, "Summary": Briefly summarize this operation}\\
  \midrule
    \textbf{User during the first operation}\\
    \hline
    The first image is the screen recording, in which the tasks are completed: \{$I_v$\}\\
    \\
    The second image is the screenshot of the current device, in which you need to complete the following tasks: \{$I_u$\}\\
    \\
    Note: You need to refer to the operation path in the video more than relying on your own operation experience. Because you may make mistakes.\\
    \\
    Note: You need to refer to the operation path in the video more than relying on your own operation experience. Because you may make mistakes."\\
    <image: $V_w$><image: $D_i$>\\
  \midrule
    \textbf{User during subsequent operations}\\
    \hline
    The first image is the screen recording, in which the tasks are completed: \{$I_v$\}\\
    \\
    The second image is the screenshot of the current device, in which you need to complete the following tasks: \{$I_u$\}\\
    \\
    Here is your operation history:\\
    Step-1: \{operation 1\}\\
    Step-2: \{operation 2\}\\
    ......\\
    Step-n: \{operation n\}\\
    \\
    Note: If the operation history and current device can infer that the task has been completed, use Done.\\
    \\
    Note: You need to refer to the operation path in the video more than relying on your own operation experience. Because you may make mistakes."\\
    <image: $V_w$><image: $D_i$>\\
    \bottomrule
    \hline
	\end{tabular}
	}
    \caption{The prompt for decision agent.}
	\label{tb:prompt_decision}
\end{table*}

\begin{table*}[t]
	\centering
	\renewcommand{\arraystretch}{1}
	\setlength{\tabcolsep}{8pt}
	\scalebox{0.9}{
	\begin{tabular}{p{17cm}}
        \hline
		\toprule
  \textbf{System}\\
  \hline
    You are a mobile phone operation assistant. I will provide you with two images. The first image is a long picture of key frames from a mobile phone operation video, which shows a correct operation trajectory to complete the task: \{$I_v$\}. The second image is two screenshots before and after an operation from the user. The user want to complete the task: \{$I_u$\}. Please note that these two images are not necessarily the complete operation trajectories, they may only be part of the continuous operation.\\
    \\
    Although the task shown in the video may not be exactly the same as the task the user needs to complete, there is a strong correlation between the two. So the user is referring to the operation in the video to complete this task.\\
    \\
    Now you need to determine which frame of the video the user is in after the device is operated. You need to use a number to represent it. If the device is in the state between two frames, the previous frame is output.\\
    If the device is not in any frame of the video, please output the number 0 to indicate an operation error and generate an error cause analysis.\\
    \\
    You need to output in the following json format:\\
    \{"Thought": Your thought of current question, "Frame": a number, "Analysis": If Frame is 0, generate an error cause analysis, otherwise output null, "Need\_Back": If Frame is 0, you need to think about how to get back on track. If you need to return to the previous page, please output true. If you need to continue to perform an operation on the current page to get back on track, please output false. If Frame is not 0, please output False directly.\}\\
  \midrule
    \textbf{User}\\
    \hline
    Here are the video and operation:\\
    <image: $V_w$><image: $D_i$>\\
    \bottomrule
    \hline
	\end{tabular}
	}
    \caption{The prompt for video agent.}
	\label{tb:prompt_video}
\end{table*}

\begin{table*}[h]
    \centering
    \renewcommand{\arraystretch}{1.2}
    \scalebox{0.9}{
    \begin{tabular}{l|p{15cm}}
        \hline
		\toprule
        \textbf{Applications} & \textbf{Task Name}\\
        \hline
		\toprule
        Expense&ExpenseAddMultiple, ExpenseAddMultipleFromGallery, ExpenseAddMultipleFromMarkor, ExpenseAddSingle, ExpenseDeleteDuplicates, ExpenseDeleteDuplicates2, ExpenseDeleteMultiple, ExpenseDeleteMultiple2, ExpenseDeleteSingle\\
        \hline
        Markor&MarkorAddNoteHeader, MarkorChangeNoteContent, MarkorCreateFolder, MarkorCreateNote, MarkorCreateNoteAndSms, MarkorCreateNoteFromClipboard, MarkorDeleteAllNotes, MarkorDeleteNewestNote, MarkorDeleteNote, MarkorEditNote, MarkorMergeNotes, MarkorMoveNote, MarkorTranscribeReceipt, MarkorTranscribeVideo\\
        \hline
        Recipe&RecipeAddMultipleRecipes, RecipeAddMultipleRecipesFromImage, RecipeAddMultipleRecipesFromMarkor, RecipeAddMultipleRecipesFromMarkor2, RecipeAddSingleRecipe, RecipeDeleteDuplicateRecipes, RecipeDeleteDuplicateRecipes2, RecipeDeleteDuplicateRecipes3, RecipeDeleteMultipleRecipes, RecipeDeleteMultipleRecipesWithConstraint, RecipeDeleteMultipleRecipesWithNoise, RecipeDeleteSingleRecipe, RecipeDeleteSingleWithRecipeWithNoise\\
        \hline
        SportsTracker&SportsTrackerActivitiesCountForWeek, SportsTrackerActivitiesOnDate, SportsTrackerActivityDuration, SportsTrackerLongestDistanceActivity, SportsTrackerTotalDistanceForCategoryOverInterval, SportsTrackerTotalDurationForCategoryThisWeek\\
        \hline
        Tasks&TasksCompletedTasksForDate, TasksDueNextWeek, TasksDueOnDate, TasksHighPriorityTasks, TasksHighPriorityTasksDueOnDate, TasksIncompleteTasksOnDate\\
        \bottomrule
        \hline
    \end{tabular}
    }
    \caption{Tasks in AndroidWorld-Knowledge.}
    \label{benchmark2}
\end{table*}

\begin{table*}[h]
    \centering
    \renewcommand{\arraystretch}{1.25}
    \scalebox{0.85}{
    \begin{tabular}{l|l|p{7cm}|p{7cm}}
        \hline
        \toprule
        APP & \textbf{Level} & \textbf{Video Instruction \& Video-Aligned User Instruction} & \textbf{Video-Misaligned User Instruction} \\
        \hline
        Phone & Basic & Help me dial 123. & Help me dial 321. \\
        & Normal & Please turn on the call recording for me. & Please view all call recording for me. \\
        & Advanced & Help me add the mobile number 1234567890 to the blacklist. & Help me add the mobile number 9876543210 to the whitelist. \\
        \hline
        Messages & Basic & Help me set up messages and notifications to be displayed together in Messages. & Help me set up messages and notifications not to be displayed together in Messages. \\
        & Normal & Please send a message to 123456 with text "Hello" & Please send a message to 9876543210 with text "Goodbye". \\
        & Advanced & Send a message to 123456 with my current location information. & Send a message to 987654 with my contact card. \\
        \hline
        Setting & Basic & Help me turn off the auto brightness in Setting. & Help me turn on the auto brightness in Setting. \\
        & Normal & Help me turn off the status bar network speed display. & Help me turn off the status bar NFC display. \\
        & Advanced & Help me open three-finger screenshots. & Help me open three-finger touch and hold. \\
        \hline
        Photo & Basic & Help me turn on the shared albums setting in Photos. & Help me turn off the shared albums setting in Photos. \\
        & Normal & Help me clear recently deleted photos. & Help me restore recently deleted photos. \\
        & Advanced & Help me set up not to record location when taking photos. & Help me set up not to record properties when taking photos. \\
        \hline
        Manager & Basic & Help me turn on the App cleaner reminder in Phone Manager. & Help me turn off the App cleaner reminder in Phone Manager. \\
        & Normal & Help me turn on the automatic phone call for help. & Help me turn on the automatic phone call for help and countdown sound. \\
        & Advanced & Help me clean up QQ's storage. & Help me clean up WhatsApp's storage. \\
        \hline
        Recorder & Basic & Help me start recording. & Help me stop recording. \\
        & Normal & Help me change the audio format of my recording. & Help me turn on the cloud recording. \\
        & Advanced & Help me show recently deleted recordings. & Help me show call recordings. \\
        \hline
        Files & Basic & Help me view photos in My Files. & Help me view videos in My Files. \\
        & Normal & Help me create a new tag named "test". & Help me create a new tag named "mobile". \\
        & Advanced & Help me turn on the option to show hidden files. & Help me turn off the option to show hidden files. \\
        \hline
        Clock & Basic & Help me start stopwatch in Clock. &  Help me reset stopwatch in Clock. \\
        & Normal & Help me set the gesture to turn off the alarm to swipe up. & Help me set the gesture to turn off the alarm to press button. \\
        & Advanced & Help me delete the last city of the current world clock and add London. & Help me delete the first city of the current world clock and add New York. \\
        \hline
        Weather & Basic & Help me turn on the meteorological alert setting in Weather. &  Help me turn off the meteorological alert setting in Weather. \\
        & Normal & Help me turn on the rain reminder. & Help me turn off the rain reminder. \\
        & Advanced & Help me turn on the UV intensity display and view the UV intensity at your current location. & Help me turn on the Sunset display and view the sunset at your current location. \\
        \hline
        Calendar & Basic & Help me turn on fixed time zone setting in Calendar. &  Help me turn off fixed time zone setting in Calendar. \\
        & Normal & Help me turn on calendar meeting reminders. & Help me turn on fixed time zone. \\
        & Advanced & Help me subscribe to horoscope and choose Aries. & Help me subscribe to today in history. \\
        \bottomrule
        \hline
    \end{tabular}
    }
    \caption{Tasks in Mobile-knowledge.}
    \label{benchmark}
    \label{tab:benchmark}
\end{table*}

\end{document}